\documentclass{article}
\pdfoutput=1

\usepackage[preprint]{neurips_data_2024}

\usepackage[utf8]{inputenc} 
\usepackage[T1]{fontenc}    
\usepackage{hyperref}       
\usepackage{url}            
\usepackage{booktabs}       
\usepackage{amsfonts}       
\usepackage{nicefrac}       
\usepackage{microtype}      
\usepackage{xcolor}         
\usepackage{amsmath}
\usepackage{amssymb}
\usepackage{threeparttable}
\usepackage{CJKutf8}
\usepackage{multirow}
\usepackage{graphicx}
\usepackage{array}
\usepackage{enumitem}

\title{HARMONIC: Harnessing LLMs for Tabular Data Synthesis and Privacy Protection}

\author{%
  Yuxin Wang \\
  Sichuan University\\
  Chengdu, China \\
  \texttt{wangyuxin1st@gmail.com} \\
  \And
  Duanyu Feng \\
  Sichuan University\\
  Chengdu, China \\
  \texttt{fengduanyuscu@stu.scu.edu.cn} \\
  \And
  Yongfu Dai\\
  Sichuan University \\
  Chengdu, China \\
  \texttt{wal.daishen@gmail.com} \\
  \And
  Zhengyu Chen \\
  Wuhan University \\
  Wuhan, China \\
  \texttt{2019302120293@whu.edu.cn} \\
  \And
  Jimin Huang \\
  The Fin AI \\
  Singapore \\
  \texttt{jimin.huang@thefin.ai} \\
  \And
  Sophia Ananiadou \\
  The University of Manchester\\
  Manchester, UK \\
  \texttt{sophia.ananiadou@manchester.ac.uk} \\
  \And
  Qianqian Xie$^*$ \\
  The Fin AI \\
  Singapore \\
  \texttt{qianqian.xie@thefin.ai} \\
  \And
  Hao Wang\thanks{Co-Corresponding Author.} \\
  Sichuan University \\
  Chengdu, China \\
  \texttt{wangh@scu.edu.cn} \\
}

\begin{document}

\maketitle

\begin{abstract}
Data serves as the fundamental foundation for advancing deep learning, particularly tabular data presented in a structured format, which is highly conducive to modeling.
However, even in the era of LLM, obtaining tabular data from sensitive domains remains a challenge due to privacy or copyright concerns. 
Hence, exploring how to effectively use models like LLMs to generate realistic and privacy-preserving synthetic tabular data is urgent.
In this paper, we take a step forward to explore LLMs for tabular data synthesis and privacy protection, by introducing a new framework HARMONIC for tabular data generation and evaluation.
In the tabular data generation of our framework, unlike previous small-scale LLM-based methods that rely on continued pre-training, we explore the larger-scale LLMs with fine-tuning to generate tabular data and enhance privacy.
Based on idea of the k-nearest neighbors algorithm, an instruction fine-tuning dataset is constructed to inspire LLMs to discover inter-row relationships. 
Then, with fine-tuning, LLMs are trained to remember the format and connections of the data rather than the data itself, which reduces the risk of privacy leakage.
In the evaluation part of our framework, we develop specific privacy risk metrics DLT for LLM synthetic data generation, as well as performance evaluation metrics LLE for downstream LLM tasks. 
Our experiments find that this tabular data generation framework achieves equivalent performance to existing methods with better privacy, which also demonstrates our evaluation framework for the effectiveness of synthetic data and privacy risks in LLM scenarios.
\end{abstract}

\section{Introduction}
In the age of deep learning, tabular data is a predominant data format and a key element for building more effective algorithms to solve specific applications in various fields \cite{fang2024large,lu2024large}.
However, in many sensitive domains such as business \cite{mottini2018airline}, healthcare \cite{chen2021synthetic}, and governmental operations \cite{jeong2022customs}, there are significant limitations on the acquisition and use of tabular data.
Tabular data in these domains involves personal privacy, business secrets, or state secrets. The collection and use of such data are strictly regulated by laws and regulations, and compliance with relevant data protection requirements is necessary. Unauthorized use or disclosure may result in serious privacy infringement or business losses.
Therefore, generating data that ensures the effectiveness in modeling these data while preserving privacy in tabular data synthesis has always been a critical research area \cite{zhao2023tabula,liu2024scaling,carey2024dp}.

Traditionally, tabular data synthesis relied on methods like GANs \cite{xu2019modeling, zhao2021ctab,wen2022causal}, VAEs \cite{tazwartab,apellaniz2024improved}, and Diffusion Models \cite{kotelnikov2023tabddpm, zhang2023mixed,liu2024controllable,sattarov2023findiff}. These techniques, built on mathematical foundations and complex frameworks, significantly advanced the field.
However, the rise of Large Language Models (LLMs) with their impressive ability to generate realistic data has shifted the paradigm. Methods like GReaT \cite{borisov2022language} and TabuLa \cite{zhao2023tabula} leverage LLMs for faster synthesis by converting tables to natural language and predicting the next data token. They often utilize smaller pre-trained models like GPT-2 \cite{sanh2019distilbert} for efficiency.
Despite their advantages, LLMs introduce significant privacy concerns \cite{yan2024protecting,mandal2024initial}. These models may potentially leak sensitive information from the training data they are exposed to. Therefore, a crucial area of exploration lies in developing strategies to mitigate these privacy risks while harnessing the power of LLMs for tabular data synthesis.

To \underline{Har}ness LL\underline{M}s f\underline{O}r Tabular Data Sy\underline{N}thesis and Pr\underline{I}vacy Prote\underline{C}tion, we develop a new framework, HARMONIC\footnote{\url{https://github.com/Wendy619/HARMONIC}.}\label{github}, with tabular data generation and evaluation on LLMs. 
For the tabular data generation framework, we use existing larger-scale LLMs to leverage their understanding abilities for generating tabular data while ensuring privacy.
It is based on the idea of k-nearest neighbor classification (kNN) \cite{cover1967nearest}, which lets the LLMs see the relationship between multiple similar rows and construct the structural tabular synthetic data format. 
With this format, we obtain the instruction-tuning datasets that retain more structural information for LLMs to enhance the ability to generate synthetic data through fine-tuning, while avoiding the forced memorization of data with pre-training.
Meanwhile, to comprehensively assess the effectiveness and privacy of synthetic data generated by LLMs, our framework introduces two novel metrics: DLT (Data Leakage Test) and LLE (LLM Efficiency). DLT quantifies the privacy risk of the synthesized data by LLMs. Conversely, LLE evaluates the effectiveness of the synthetic data in downstream LLM tasks.
The evaluation of the effectiveness of downstream LLM tasks is based on the increasing application of LLMs in various fields. It is important to note that machine learning-based evaluations are no longer sufficient.

Using our evaluation framework, we assess four datasets commonly used for classification tasks in tabular data synthesis. The results show that synthetic data generated with HARMONIC performs comparably to existing methods in machine learning and excels in downstream tasks and privacy assessments in LLMs. Crucially, HARMONIC's evaluation suggests that traditional synthetic data methods may be unsuitable for downstream LLM tasks and that pretraining-based synthetic data poses significant privacy risks.


The main contributions of this study can be summarized as follows: 1) We recognize that it is crucial to not only focus on the strong data generation ability of LLM in this era, but also pay attention to the potential privacy risks it may bring. 2) We develop a framework, HARMONIC, for synthesizing tabular data based on LLM. The framework aims to minimize the risk of data leakage while ensuring the effectiveness of data synthesis using LLM. 3) Under the HARMONIC framework, a set of metrics is proposed for the effectiveness in downstream LLMs tasks and privacy risk evaluation of synthetic tabular data.

\section{Related work}

\textbf{Tabular Data Synthesis}.
Prior to the rise of Large Language Models (LLMs), synthetic tabular data generation primarily relied on machine learning or classical neural network frameworks. These methods can be broadly categorized into three groups: Simple Augmentation, Generative Adversarial Networks (GANs), and Diffusion Models. Techniques like SMOTE \cite{chawla2002smote} exemplify Simple Augmentation, leveraging linear interpolation for data resampling. While effective for structured data, SMOTE overlooks semantic information. Building on GANs, CTGAN \cite{xu2019modeling} introduces a conditional generator and adapts a Variational Autoencoder (VAE) for tabular data (TVAE). CTAB-GAN \cite{zhao2021ctab} tackles data imbalance and long-tail issues. TabDDPM \cite{kotelnikov2023tabddpm} serves as a prominent benchmark for Diffusion-based methods, with TABSYN \cite{zhang2023mixed} offering faster synthesis compared to other such techniques. However, most of these methods utilize one-hot encoding for categorical data, which can exacerbate the "curse of dimensionality" for high-cardinality variables and fail to capture contextual information \cite{borisov2022language, zhao2023tabula}.

LLMs have emerged as a compelling approach for synthetic data generation due to their exceptional capabilities in producing high-quality, human-like data. LLM-based methods commonly employ a pre-training paradigm. Real tabular data is converted into text format and fed into the LLM for learning. GreaT \cite{borisov2022language} exemplifies this approach, converting each tabular feature into the format "X is Y" and feeding the text into GPT-2 \cite{sanh2019distilbert} for fine-tuning.  Tabula \cite{zhao2023tabula} introduces a tabular data synthesizer leveraging an LLM framework without pre-trained weights. It prioritizes faster training speed by simplifying token sequences to "X Y". REaLTabFormer \cite{solatorio2023realtabformer} presents a transformer-based framework for generating both non-relational and relational tabular data. It treats each tabular sample as a sequence with dependencies, akin to a sentence, learning conditional distributions to sequentially generate complete samples.
While LLM-based methods often outperform machine learning approaches due to their ability to leverage contextual information in text entries, limitations exist. Processing table data entry-by-entry hinders LLMs from fully exploiting relational information between samples. Furthermore, inherent security risks associated with data leakage plague LLMs \cite{yan2024protecting}. Pre-training-like fine-tuning can make them vulnerable, potentially allowing an attacker with knowledge of one or two feature values in a real entry to retrieve the entire real data record.

\textbf{Tabular Data Synthesis Evaluate}.
Existing evaluation methods for synthetic data, such as the MLE benchmarking system proposed by Xu et al. \cite{xu2019modeling}, primarily focus on assessing its performance as training data for machine learning models. However, as Kotelnikov et al. \cite{kotelnikov2023tabddpm} argue, relying on weak classifiers for evaluation becomes outdated in light of the capabilities of advanced models like CatBoost \cite{prokhorenkova2018catboost}. This underscores the need for more sophisticated evaluation techniques, especially considering the widespread adoption of LLMs in downstream applications \cite{feng2023empowering}.

Current privacy metrics for synthetic data, such as Distance to Closest Record (DCR) \cite{zhao2021ctab} and the NewRowSynthesis metric from SDMetrics \cite{sdmetrics}, solely analyze the distance between synthetic data and real data. While these distance-based approaches provide valuable insights, they fall short when dealing with Large Language Models (LLMs). 
LLMs are particularly susceptible to data leakage due to their complex nature and training on massive datasets \cite{yan2024protecting}. However, existing privacy metrics based solely on tabular data feature distances fail to capture the unique learning and inference mechanisms of LLMs, which operate at the semantic and generative probability levels of embeddings. Consequently, these methods lack intuitive indicators of privacy leakage specific to LLMs \cite{wang2024pandora}.

\section{Our Framework: HARMONIC}

This chapter delves into the HARMONIC framework for tabular data synthesis powered by LLMs, encompassing both generation and evaluation modules.

\subsection{Synthetic Tabular Data Generation Framework}
\begin{figure*}[!htb]
\vspace{-3mm}
\centering
\includegraphics[width=0.8\columnwidth]{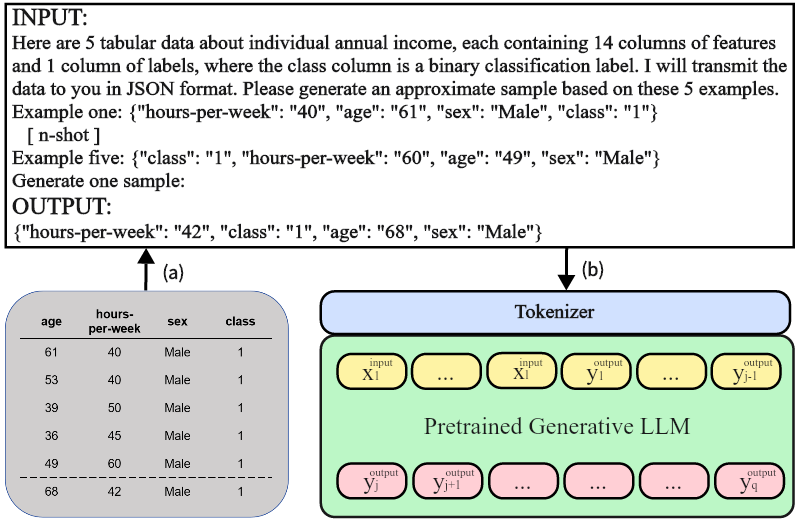} 
\vspace{-4mm}
\caption{After applying the kNN algorithm to the original table, we obtain $n$ sets of $k+1$ data points. Each set is structured according to the template shown in the gray table at the bottom left. These datasets are then encoded into a single instruction using text encoding, with the features of each table data shuffled, as shown in the white box above (a). Finally, the encoded fine-tuning dataset is input into the pre-trained LLM for fine-tuning (b).}
\label{ft}
\vspace{-4.5mm}
\end{figure*}

\begin{figure*}[!htb]
\centering
\includegraphics[width=1\columnwidth]{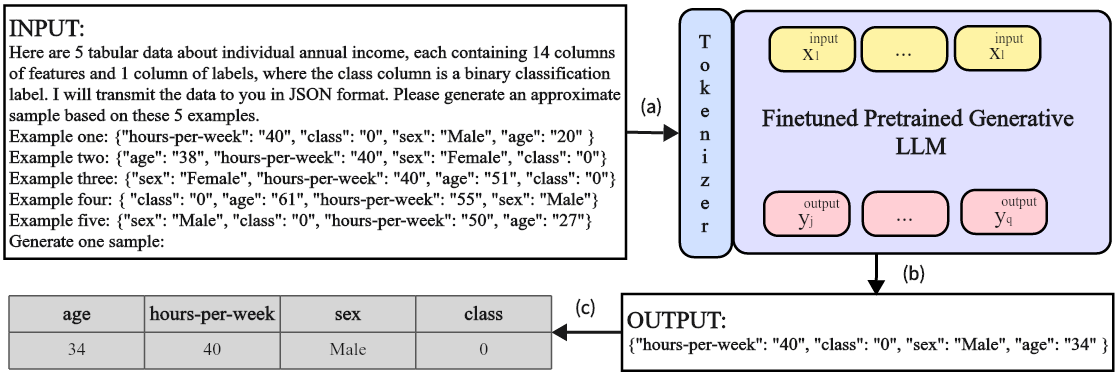} 
\vspace{-4mm}
\caption{The sampling step involves inputting a prompt, shown within the white box in the upper left corner (a), into the fine-tuned pretrained LLM. This results in a textual output (b), which is then converted into a table using pattern matching (c).}
\label{sm}
 \vspace{-3mm}
\end{figure*}

In this section, we present our approach to fine-tuning pre-trained LLMs for the generation of synthetic tabular data, including three key stages:
 (1) \textbf{Construct instruction dataset}: Construct an instruction fine-tuning dataset designed to fine-tune the generator model and a prompt dataset to facilitate data generation.
 (2) \textbf{Instruction tuning}: Fed the instruction fine-tuning dataset into a pre-trained LLM for fine-tuning, as illustrated in Figure \ref{ft}; 
 (3) \textbf{Sampling}: Synthetic tabular data is generated by sampling from the fine-tuned LLM, with the sampling process depicted in Figure \ref{sm}. Below, we will provide a comprehensive description of the entire process, encompassing the construction of the instruction dataset, model fine-tuning, and the implementation of sampling. 

\subsubsection{Construct Instruction Dataset}
\label{data pre}
\textbf{Construct a fine-tuning dataset using kNN.}
Our approach leverages kNN to enable LLMs to generate synthetic data resembling limited real data. This is expected to use the in-context learning ability of LLM (few-shots)  to mine information from the most relevant table samples.

Specifically, this process involves finding the $k$ nearest neighbors for each training sample, creating a set of $k+1$ data points. 
To improve the quality of the generated synthetic data, a filtering step is necessary. Specifically, for each set of $k+1$ data, if more than half of the input data have labels that are different from that of the single corresponding data point, this $k+1$ data is discarded. Ultimately, this filtering process yields $n$ sets of $k+1$ data. A constant value of $k = 5$ was used throughout our experimental setup.

\textbf{Data format engineering.} Since LLMs are designed as sequence-to-sequence models, feeding tabular data into an LLM requires converting the structured data into a textual format. A straightforward approach would be to directly input a programming language readable data structure, such as Pandas DataFrame Loader for Python, line-separated JSON-file format, HTML code reflecting tables, etc. \cite{fang2024large}
In our method, each table entry is converted into JSON dictionary format, preserving the original table structure and enabling the model to understand the semantics of each value.

For a table entry $s_i$ with feature names $f_1, f_2, \ldots, f_m$, where the value of its $j$-th feature is $v_{i,j}$, the JSON-formatted data $t_i$ corresponding to the table entry $s_i$ is defined as follows:
\begin{align}\label{AA}
t_{i,j} = [f_j:v_{i,j}]\qquad&\forall i\in\{1,\ldots,n(k+1)\}, j\in\{1,\ldots,m\},\\
\textbf{t}_i=\{t_{i,1},t_{i,2},\ldots,t_{i,m}\}\qquad&\forall i\in\{1,\ldots,n(k+1)\},
\end{align}
We concatenate $k$ JSON-formatted data entries sequentially, incorporating prompts to elucidate the fine-tuning task and contextualize the data. The label JSON-formatted data entry serves as the reference answer.
In addition, when converting a tabular feature vector into a sequence using the text encoding scheme, we inadvertently introduce pseudo-positional information into the transformed tabular data sample. However, there is no inherent spatial ordering among features in tabular datasets\cite{borisov2110deep}. To restore feature order independence, we randomly shuffle the order of features within each complete JSON-formatted data entry $\textbf{t}_i$ using a permutation. This operation results in a new sequence where the order of features is randomized, ensuring that the model learns to be invariant to feature order. Therefore, a template for this instruction fine-tuning dataset is shown as Figure \ref{ft} \footnote{For illustrative examples, please refer to Appendix \ref{data instance}.}. 

\textbf{Construct prompt dataset for generation.}
To generate synthetic data, we need to construct a prompt dataset consistent in format with the fine-tuning dataset. There are three key differences between the prompt dataset and the fine-tuning dataset: (1) The prompts used for generating data remain consistent with those used during fine-tuning, with the exception of the $\mathrm{OUTPUT}$ field, as the output is the synthetic data that the model needs to generate without reference answer. (2) Each set of $k$ real data points in the prompt dataset is randomly resampled from the real data, unlike in the fine-tuning dataset, preventing the model from reproducing the original real data, and it does not require filtering operations. (3) The size of the prompt dataset should be larger than the number of synthetic data samples required, as each prompt can generate only one piece of synthetic data.




\subsubsection{Instruction Tuning}
We then fine-tune the LLM for the synthetic data generation task using the instruction dataset we constructed. Unlike pretraining-based LLMs for this task, our aim is to prevent the LLM from memorizing the original tabular data in the dataset. 
After tokenizing our instruction dataset, the resulting token embeddings of one sample for the $\mathrm{INPUT}$ and $\mathrm{OUTPUT}$ are denoted as $\mathrm{emb(X)}=(x_1,\ldots,x_l)$ and $\mathrm{emb(Y)}=(y_1,\ldots,y_q) $, respectively. Here, $l$ and $q$ represent the lengths of the $\mathrm{INPUT}$ and $\mathrm{OUTPUT}$, respectively. Therefore, the objective of our fine-tuning strategy is to maximize the probability of generating the correct output sequence given a prompt describing the task and $k$ input real data points. This objective function is formulated as:
\begin{align}
\begin{aligned}
p(\textbf{ft})=p(\mathrm{emb(Y)}|\mathrm{emb(X)})=p(y_1,\ldots,y_q|x_1,\ldots,x_l)=\prod_{j=1}^qp(y|x_1,\ldots,x_l,y_1,\ldots,y_j)\label{BB}
\end{aligned}
\end{align}
The LLM is trained by optimizing the parameters to maximize the probability $\prod_{\textbf{ft}\in FT} p(\textbf{ft})$, which only compute the loss of $\mathrm{OUTPUT}$ and avoids learn the real data in the $\mathrm{INPUT}$ to protect privacy.

\subsubsection{Sampling}

We denote the fine-tuned LLM as the generator $\textbf{G}$. Each data point in the prompt dataset is fed into $\textbf{G}$, yielding the distribution of subsequent tokens conditioned on the known input sequence. To generate the next token with more diversity and protect privacy, we adopt a weighted sampling strategy incorporating a temperature coefficient $T$. We set the default temperature coefficient $T$ to 0.7. After generation, we utilize pattern-matching algorithms, as described in \cite{aho1990algorithms}, to reconvert the generated textual feature representations into a dataframe format, resulting in the final synthetic tabular dataset.

\subsection{Synthetic Tabular Data Evaluation Framework}

We introduce two new metrics to evaluate the quality and privacy of synthetic data for LLM-based synthesis methods: LLM Efficacy (LLE) and Data Leakage Test (DLT).
\subsubsection{LLE: LLM Efficacy}


With the development of LLMs, we believe that evaluating the quality of synthetic data using weak classifiers is losing its practical value and credibility. More and more pepole are concerned with the performance of synthetic data as a training set for state-of-the-art methods \cite{kotelnikov2023tabddpm}. Recent research exploring the application of LLMs to tabular data processing has yielded significant advancements, with potential to rival or even surpass state-of-the-art machine learning approaches \cite{yan2024making}. Therefore, we propose using synthetic data to fine-tune a pretrained LLM and then evaluate the fine-tuned LLM on the real test set. We refer to this as \textbf{LLM Efficacy} (\textbf{LLE}). We choose LLaMA-2-7b-chat \cite{touvron2023llama} as the base model to compute the \textbf{LLE}.


\subsubsection{DLT: Data Leakage Test}


The metrics Distance to Closest Record (DCR) \cite{zhao2021ctab} and SDMetrics \cite{sdmetrics} focus on measuring the "distance" between synthetic data and real data, without taking into account the extent to which the generator itself leaks data. Research indicates that LLMs are susceptible to data leakage issues to varying degrees \cite{yan2024protecting}. Attacks on LLMs of synthetic data generator can potentially extract complete training data, leading to severe privacy breaches. To address this, we propose a new metric for quantifying privacy protection, named the \textbf{Data Leakage Test (DLT)}, inspired by the work of Skywork \cite{wei2023skywork}. This metric measures the extent to which a generator leaks real data, thereby reflecting the privacy level of the synthetic data. The $\mathrm{DLT}$ computes the perplexity of the generator on both synthetic and real data to determine its data generation tendencies.

To compute the $\mathrm{DLT}$, firstly, we feed the training data into the generator to calculate the ppl(perplexity) for each sample, then average these scores to determine the ppl on the training data, referred to as <ppl-on-train>. Then, we feed synthetic data into the generator and obtain the ppl on synthetic data, referred to as <ppl-on-syn>. The DLT value is computed by subtracting <ppl-on-syn> from <ppl-on-train>. A larger DLT value indicates better privacy preservation of the original data by the generator, whereas a smaller value indicates weaker privacy preservation. The computation formula of DLT is as below, where the $P(x)$ denotes the probability of generating a sentence.
\begin{align}
\mathrm{DLT} &= \mathrm{PPL(D_{test})} - \mathrm{PPL(D_{train})}\label{EE}
\end{align}
\begin{align}
\begin{aligned}
\mathrm{PPL(D_{split})}=\frac{1}{|\mathrm{D_{split}}|}\sum_{x\in \mathrm{D_{split}}}P(x)^{-\frac{1}{N}} = \frac{1}{|\mathrm{D_{split}}|}\sum_{x\in \mathrm{D_{split}}}2^{\mathrm{Cross-Entropy(x)}}\label{FF}
\end{aligned}
\end{align}

\section{Experiment}




In this section, we select four real-world datasets to compare the performance of HARMONIC with various types of data synthesis methods. The comparison is conducted from two perspectives: the effectiveness of the synthesized data and its privacy.

\subsection{Experimental Setup}
\label{exp set}
\textbf{Datasets.}
To evaluate the proposed method, we utilized four real-world datasets from various domains, namely GM (German \cite{misc_statlog_(german_credit_data)_144}), AD (Adult Income \cite{kohavi1996scaling}), DI (Diabetes)\footnote{\url{https://www.openml.org/search?type=data&sort=runs&id=37}}, BU (Buddy)\footnote{\url{https://www.kaggle.com/datasets/akash14/adopt-a-buddy}}, which are all open source datasets and don't contain any personal information such as names, phone numbers, addresses, or other sensitive data.
These datasets differ in size, feature types, and the number of features, ranging from fewer than 1,000 to tens of thousands of samples. Some datasets include only numerical features, while others contain both numerical and categorical features. We divided each dataset into training, validation, and test sets in approximately a 7:1:2 ratio. All models were trained on the same training data samples.

\textbf{Baselines.}
There are numerous synthetic methods for generating tabular data. Based on the classification approach discussed previously, we selected the most representative methods as our baselines.
SMOTE \cite{chawla2002smote} is a simple interpolation method proposed for oversampling minority classes and can also be used for generating synthetic data. TVAE \cite{xu2019modeling} is a state-of-the-art method for tabular data generation based on VAE. CTABGAN \cite{zhao2021ctab} is a GAN-based model that performs exceptionally well across a diverse set of benchmarks. TabDDPM \cite{kotelnikov2023tabddpm} serves as a famous benchmark for Diffusion-based Methods. TABSYN \cite{zhang2023mixed} achieves faster synthesis compared to other diffusion-based techniques. GReaT \cite{borisov2022language} and REaLTabFormer \cite{solatorio2023realtabformer} are SOTA tabular data synthesizers based on LLMs, to be precise, both are based on GPT-2 \cite{sanh2019distilbert}. The code for these methods can be found on GitHub.

\textbf{Metrics.}
For the effectiveness of synthetic data, we evaluate it using our proposed LLE metric. Specifically, we convert the synthetic tabular data into the text format required for specific classification tasks, then feed this data into a pre-trained LLM for fine-tuning. The fine-tuned model is then tested using a test set that has been similarly converted into the corresponding text format, and the weighted average F1 score is obtained. We also fine-tune the pre-trained LLM using the training set of real data and evaluate it on the test set. The synthetic data is considered to have practical value if the performance of the fine-tuned model using synthetic data is on par with or better than that using real data.

For the privacy of synthetic data, we use three different metrics to evaluate privacy: DCR \cite{zhao2021ctab} and NRS($\mathrm{NewRowSynthesis}$) \cite{sdmetrics}, and our proposed DLT metric. All three metrics are positively correlated with privacy, meaning that higher values indicate stronger privacy.

\textbf{Implementation Details.} \footnote{For detailed configurations of the fine-tuning, please refer to Appendix \ref{Exp}.}
Our approach allows for the selection of any pre-trained generative LLM that supports fine-tuning, such as GPT-2 \cite{sanh2019distilbert}, LLaMA-2-7b-chat \cite{touvron2023llama}, Mistral \cite{jiang2023mistral}, etc., as the base model. By default, our method opts for LLaMA-2-7b-chat \cite{touvron2023llama} as the base model due to its rich pre-training corpus, resulting in a stronger language understanding capability compared to GPT-2 \cite{sanh2019distilbert}. This enables LLaMA-2-7b-chat \cite{touvron2023llama} to learn fine-tuning tasks more efficiently. However, users also have the flexibility to switch base models according to their specific requirements. Considering the time cost of the entire experiment, we choose lora \cite{hu2021lora} efficient fine-tuning instead of full parameter adjustment. 

\subsection{The Effectiveness of Synthetic Data}
\textbf{Our experimental results offer compelling evidence that the synthetic data generated by our method can effectively serve as a substitute for real data in downstream tasks.} This finding aligns with the growing recognition that traditional Machine Learning Efficacy (MLE) metrics may not be well-suited for evaluating the effectiveness of synthetic data used with modern LLMs. Relying solely on MLE metrics can be misleading when evaluating LLMs, potentially leading to inaccurate conclusions.

\vspace{-3mm}
\renewcommand{\arraystretch}{1.8}
\begin{table*}[!htb]
\centering
\caption{The results for effectiveness. The best results are marked in bold, the second-best results are underlined. All results are averages over 3 trials with different random seeds.}
\vspace{1mm}
\label{tab:LLE}
\resizebox{0.999\textwidth}{!}{
    \begin{tabular}{
    >{\centering\arraybackslash}m{1.2cm}
    >{\centering\arraybackslash}m{1.2cm}
    >{\centering\arraybackslash}m{1.2cm}
    >{\centering\arraybackslash}m{1.5cm}
    >{\centering\arraybackslash}m{1.5cm}
    >{\centering\arraybackslash}m{1.2cm}
    >{\centering\arraybackslash}m{1.2cm}
    >{\centering\arraybackslash}m{1.3cm}
    >{\centering\arraybackslash}m{1.3cm}
    >{\centering\arraybackslash}m{1.2cm}
    >{\centering\arraybackslash}m{1.2cm}}
    \toprule
    Dataset & Metric & Original & HARMONIC & SMOTE & TVAE & CTAB & TabDDPM & TABSYN & GReaT & RTF\\
    \midrule
    \multirow{2}{*}{GM} & MLE & $0.50_{\pm 0.00}$ & $0.55_{\pm 0.03}$ & $\underline{0.64_{\pm 0.02}}$ & $0.61_{\pm 0.02}$ & $0.57_{\pm 0.02}$ & $\underline{0.64_{\pm 0.01}}$ & $0.63_{\pm 0.02}$ & $0.44_{\pm 0.03}$ & $\textbf{0.65}_{\pm \textbf{0.01}}$ \\
     & LLE & $0.71_{\pm 0.00}$ & $0.64_{\pm 0.03}$ & $0.67_{\pm 0.04}$ & $0.69_{\pm 0.03}$ & $\underline{0.71_{\pm 0.02}}$ & $0.67_{\pm 0.05}$ & $\textbf{0.72}_{\pm \textbf{0.02}}$ & $0.55_{\pm 0.11}$ & $0.69_{\pm 0.03}$\\
     \hline
    \multirow{2}{*}{AD} & MLE & $0.61_{\pm 0.00}$ & $0.67_{\pm 0.02}$ & $\underline{0.75_{\pm 0.00}}$ & $0.74_{\pm 0.00}$ & $0.73_{\pm 0.01}$ & $0.74_{\pm 0.00}$ & $0.73_{\pm 0.01}$ & $0.73_{\pm 0.01}$ & $\textbf{0.76}_{\pm \textbf{0.00}}$ \\
     & LLE & $0.81_{\pm 0.00}$ & $0.80_{\pm 0.02}$ & $\underline{0.84_{\pm 0.01}}$ & $0.83_{\pm 0.01}$ & $0.83_{\pm 0.00}$ & $0.83_{\pm 0.00}$ & $0.81_{\pm 0.02}$ & $0.82_{\pm 0.02}$ & $\textbf{0.85}_{\pm \textbf{0.00}}$\\
     \hline
    \multirow{2}{*}{DI} & MLE & $0.56_{\pm 0.00}$ & $0.46_{\pm 0.02}$ & $\textbf{0.72}_{\pm \textbf{0.03}}$ & $\underline{0.71_{\pm 0.02}}$ & $0.67_{\pm 0.02}$ & $\underline{0.71_{\pm 0.02}}$ & $0.68_{\pm 0.03}$ & $0.45_{\pm 0.03}$ & $0.66_{\pm 0.03}$ \\
     & LLE & $0.70_{\pm 0.00}$ & $\underline{0.75_{\pm 0.00}}$ & $0.69_{\pm 0.04}$ & $0.72_{\pm 0.04}$ & $0.62_{\pm 0.09}$ & $0.72_{\pm 0.03}$ & $\textbf{0.77}_{\pm \textbf{0.01}}$ & $0.71_{\pm 0.03}$ & $0.70_{\pm 0.04}$\\
    \hline
    \multirow{2}{*}{BU} & MLE & $0.38_{\pm 0.00}$ & $\textbf{0.27}_{\pm \textbf{0.03}}$ & $0.25_{\pm 0.02}$ & $\textbf{0.27}_{\pm \textbf{0.03}}$ & $0.26_{\pm 0.01}$ & $\textbf{0.27}_{\pm \textbf{0.01}}$ & $0.26_{\pm 0.01}$ & $0.24_{\pm 0.03}$ & $0.26_{\pm 0.00}$ \\
     & LLE & $0.88_{\pm 0.00}$ & $0.82_{\pm 0.03}$ & $0.85_{\pm 0.04}$ & $\textbf{0.86}_{\pm \textbf{0.01}}$ & $0.82_{\pm 0.02}$ & $0.85_{\pm 0.01}$ & $\textbf{0.86}_{\pm \textbf{0.01}}$ & $0.81_{\pm 0.03}$ & $0.70_{\pm 0.14}$\\
    \bottomrule
    \end{tabular}
}
\vspace{-3mm}

\end{table*}

Therefore, we primarily base our analysis and evaluation on the LLM Efficacy (LLE) metric. This metric provides a more nuanced assessment of the quality and effectiveness of synthetic data specifically for LLM-based tasks. Table~\ref{tab:LLE} summarizes the weighted average F1 scores achieved on classification tasks using the LLaMA-2-7b-chat model. Each value in the table represents the average F1 score obtained across three independent runs of the synthetic data generation process, using different random seeds to ensure robustness.
The results presented in Table~\ref{tab:LLE} (LLE) demonstrate the effectiveness of our method. While our method surpasses the real training set on the DI dataset, its performance on the remaining three datasets falls slightly short. However, the average decrease compared to the real data benchmark is less than 5\%, which falls within an acceptable range for practical applications. Notably, even TABSYN , boasting the best overall performance among the compared methods, only outperforms the real training set on two datasets (GM and DI).
Furthermore, our method exhibits a distinct advantage in terms of stability. Compared to other prominent LLM-based methods like GReaT and RTF (REaLTabFormer), our synthetic data generation process produces results with a significantly lower standard deviation. This indicates that our method generates data with greater consistency and reliability, leading to more predictable performance in downstream LLM tasks.

In conclusion, while our method may not achieve the absolute highest performance on every dataset, the results presented in this section overwhelmingly support its potential as a viable substitute for real data. The synthetic data generated by our method demonstrates both effectiveness and stability, making it a valuable tool for various LLM-based applications.

\subsection{The Privacy of Synthetic Data}
\textbf{The experimental results demonstrate that our method prioritizes privacy in the synthetic data generation.} This is particularly beneficial in situations where disclosing real data is not feasible due to privacy concerns. In such scenarios, our synthetic data serves as a reliable and secure substitute for real data, allowing downstream tasks to proceed without compromising sensitive information. 

Table~\ref{tab:DLT} presents three key privacy metric scores to quantify the effectiveness of our method. Analyzing the results in Table~\ref{tab:DLT}, it's evident that our method surpasses or comes in a close second for almost all datasets across all three metrics. This translates to demonstrably stronger privacy protection compared to existing methods.

\vspace{-3mm}
\renewcommand{\arraystretch}{1.2}
\begin{table*}[!htb]
\centering
\caption{The results for privacy. The best results are marked in bold, the second-best results are underlined. Each dataset has three metrics, and in all cases, higher values are better.}
\vspace{1mm}
\label{tab:DLT}
\resizebox{0.9\textwidth}{!}{
    \begin{tabular}{
    >{\centering\arraybackslash}m{1.1cm}
    >{\centering\arraybackslash}m{1.1cm}
    >{\centering\arraybackslash}m{1.5cm}
    >{\centering\arraybackslash}m{1.5cm}
    >{\centering\arraybackslash}m{1.1cm}
    >{\centering\arraybackslash}m{1.1cm}
    >{\centering\arraybackslash}m{1.3cm}
    >{\centering\arraybackslash}m{1.3cm}
    >{\centering\arraybackslash}m{1.1cm}
    >{\centering\arraybackslash}m{1.1cm}}
    \toprule
    Dataset & Metric & HARMONIC & SMOTE & TVAE & CTAB & TabDDPM & TABSYN & GReaT & RTF \\
    \midrule
    \multirow{3}{*}{GM} & NRS & \textbf{1.00} & 1.00 & 1.00 & 1.00 & 1.00 & 1.00 & 1.00 & 1.00 \\
     & DCR & \textbf{8.08} & 2.77 & 4.09 & 5.36 & 2.21 & 3.98 & \underline{5.84} & 4.60 \\
     & DLT & \textbf{-0.16} & --- & --- & --- & --- & --- & \underline{-2.14} & -22.04 \\
    \hline
    \multirow{3}{*}{AD} & NRS & \textbf{1.00} & 0.95 & 1.00 & 1.00 & 1.00 & 1.00 & 1.00 & 1.00 \\
     & DCR & \textbf{2.47} & 0.16 & 0.49 & 0.82 & 0.50 & 0.86 & \underline{1.51} & 0.57 \\
     & DLT & \underline{-0.98} & --- & --- & --- & --- & --- & \textbf{-0.67} & -163.71 \\
    \hline
    \multirow{3}{*}{DI} & NRS & \textbf{1.00} & 1.00 & 1.00 & 1.00 & 1.00 & 1.00 & 1.00 & 1.00 \\
     & DCR & 0.44 & 0.28 & 0.33 & 0.72 & 0.21 & \textbf{1.37} & \underline{1.36} & 0.36 \\
     & DLT & -0.37 & --- & --- & --- & --- & --- & -0.44 & -42.46 \\
    \hline
    \multirow{3}{*}{BU} & NRS & \textbf{1.00} & 0.93 & 1.00 & 1.00 & 0.99 & 1.00 & 1.00 & 1.00 \\
     & DCR & \underline{2.52} & 0.15 & 0.66 & 0.70 & 0.18 & 1.38 & \textbf{8.30} & 0.38 \\
     & DLT & \textbf{-0.34} & --- & --- & --- & --- & --- & \underline{-2.22} & -41.13 \\
    \bottomrule
    \end{tabular}
}
\vspace{-3mm}
\end{table*}

However, the privacy benefits go beyond the quantitative metrics. The design of our method inherently offers superior security. An attacker attempting to reconstruct a single real data record would need knowledge of nearly the entire set of k real data records (typically set to 5). This includes knowing the sequence of each feature within a record and the specific order of these k samples. This significantly raises the bar for attackers compared to methods like GReaT, which exposes a vulnerability where an attacker with knowledge of just one or two feature values in a real record can potentially reconstruct the entire record.



\section{Conclusion}
\label{conclusion}
In this paper, we introduce HARMONIC, a novel framework that leverages the power of LLMs for synthesizing tabular data and privacy concerns. HARMONIC enables LLMs to capture both the internal feature relationships within individual data points and the broader connections between samples by instruction fine-tuning. Recognizing the crucial importance of privacy, we have proposed DLT specifically for detecting data privacy in LLM synthesis. Extensive evaluations across four real-world datasets for classification tasks showcase HARMONIC's ability to achieve this crucial balance of effectiveness and privacy. HARMONIC demonstrably offers robust privacy protection while preserving the effectiveness of the synthetic data. 



\textbf{Limitations}. Compared to other methods, our approach requires a longer processing time for larger LLMs. In addition, because LLMs are less sensitive to numerical data and are better suited for classification tasks rather than regression tasks. As a result, our current work focuses solely on tabular data used for classification tasks. 

{
\small
\bibliographystyle{unsrt}
\bibliography{neurips_data_2024}
}

\appendix

\newpage
\section{Datasets Details}
\label{datasets details}

\subsection{Data Source}

We list the sources of our datasets in Table \ref{data source}, all of which are obtained from publicly accessible and reputable websites.

\begin{table*}[!htb]
  \centering

  \resizebox{0.999\textwidth}{!}{
  \begin{tabular}{cl}
    \toprule
    
     Dataset & URL \\
    \midrule
     German & \url{https://archive.ics.uci.edu/dataset/144/statlog+german+credit+data} \\
    \midrule
    Adult Income & \url{https://archive.ics.uci.edu/dataset/2/adult} \\
    \midrule
    Diabetes & \url{https://www.openml.org/search?type=data&sort=runs&id=37&status=active} \\
    \midrule
    Buddy & \url{https://www.kaggle.com/datasets/akash14/adopt-a-buddy} \\
        
    \bottomrule
    \end{tabular}
}
    \caption{URLs for real-world datasets of the experiments}
    \label{data source}%
\end{table*}%

\subsection{Data Description}

Additionally, we record various statistical details for each dataset in Table \ref{data des}.

\textbf{German.} The German dataset classifies people as good or bad credit risks described by a set of attributes including status of existing checking account, duration in month, credit history, purpose and more.

\textbf{Adult Income.} The US Adult income dataset was extracted by Barry Becker from the 1994 US Census Database. The dataset consists of anonymous information such as occupation, age, native country, race, capital gain, capital loss, education, work class and more. Each row is labelled as either having a salary greater than ">50K" or "<=50K".

\textbf{Diabetes.} The Diabetes dataset originates from the National Institute of Diabetes and Digestive and Kidney Diseases. This dataset comprises medical features including the number of times pregnant, diastolic blood pressure, body mass index, age, among other variables. The label indicates whether the individual has diabetes or not.

\textbf{Buddy.} The Buddy dataset originates from the HackerEarth Machine Learning Challenge—Adopt a Buddy. The dataset consists of parameters such as: a unique ID assigned to each animal that is up for adoption, date on which they arrived at the shelter, their physical attributes such as color, length and height, among other factors. The labels in this dataset denote the breed of the animals.

\begin{table*}[!htb]
  \centering
  \resizebox{0.999\textwidth}{!}{
  \begin{tabular}{lcccccc}
    \toprule
    
    \textbf{Dataset} & \textbf{Domain} & \textbf{\# Samples} & \textbf{\# Num} & \textbf{\# Cat} & \textbf{Tasks} & \textbf{\# Classes} \\
    \midrule
    German & Financial & 1000 & 7 & 13 & Classification & 2 \\
    \midrule
    Adult Income & Social & 32561 & 6 & 8 & Classification & 2 \\
    \midrule
    Diabetes & Medical & 768 & 8 & 0 & Classification & 2 \\
    \midrule
    Buddy & Nature & 18834 & 4 & 5 & Multi-Class & 3 \\
    \bottomrule
    \end{tabular}
}
    \caption{Dataset Statistics. \# Samples denotes the number of samples in each dataset. \# Num and \# Cat columns indicate numbers of numerical and categorical features in each dataset.}
    \label{data des}%
\end{table*}%

\subsection{Data Preprocessing}

To maintain consistency in formatting, we converted all four datasets into CSV files. Additionally, the other datasets underwent the following preprocessing steps:

\textbf{German.} The original label "status" with a value of "1" was converted to "0", and the original label "status" with a value of "2" was converted to "1".

\textbf{Adult Income.} The original label "class" with a value of "<=50K" was converted to "0", and the original label "class" with a value of ">50K" was converted to "1".

\textbf{Diabetes.} The diabetes dataset was used without any additional preprocessing.

\textbf{Buddy.} The original "issue\_date" and "listing\_date," which were represented in the "date\_time" format, have been replaced with a timestamp format.

\subsection{Data Field}

The instruction fine-tunnig dataset is provided in json format and contains the following attributes. And a specific instance of INPUT and OUTPUT can be found in \ref{data instance}.

\{ \\
\hspace*{2em}\textbf{id}: [integer] The unique identifier for each instance \\
\hspace*{2em}\textbf{conversations}: [ \\
\hspace*{4em}\{ \\
\hspace*{6em}\textbf{from}: [string] "human" \\
\hspace*{6em}\textbf{value}: [string] the \textbf{INPUT} text for LLM fine-tuning\\
\hspace*{4em}\}, \\
\hspace*{4em}\{ \\
\hspace*{6em}\textbf{from}: [string] "assistant" \\
\hspace*{6em}\textbf{value}: [string] the \textbf{OUTPUT} text for LLM fine-tunnig\\
\hspace*{4em}\} \\
\hspace*{4em}] \\
\}

\subsection{Data Instance}
\label{data instance}

To illustrate the data format used for fine-tuning both the generator and downstream tasks, we present a complete data instance from the German dataset as an example, shown in Table \ref{a instance of generator} and Table \ref{a instance of downstream tasks} respectively.

\begin{table*}[!htb]
  \centering
  \resizebox{0.999\textwidth}{!}{
    \begin{tabular}{p{0.999\textwidth}}
      \toprule

       \fontsize{10}{10}\selectfont \textbf{INPUT:} Here are 5 tabular data about user credit scores, each containing 20 columns of features and 1 column of labels, where the 'status' column is a binary classification label. I will transmit the data to you in JSON format. Please generate an approximate sample based on these 5 examples.\textbackslash n Example one: \{"Present employment since": "A75", "Credit amount": "11816", "Credit history": "A30", "Purpose": "A49", "Duration in month": "45", "Other installment plans": "A143", "Age in years": "29", "Savings account/bonds": "A61", "status": "1", "foreign worker": "A201", "Number of people being liable to provide maintenance for": "1", "Number of existing credits at this bank": "2", "Installment rate in percentage of disposable income": "2", "Housing": "A151", "Property": "A123", "Present residence since": "4", "Telephone": "A191", "Other debtors / guarantors": "A101", "Job": "A173", "Status of existing checking account": "A11", "Personal status and sex": "A93"\}.\textbackslash n Example two: \{"Housing": "A151", "Personal status and sex": "A92", "Credit amount": "6416", "Job": "A173", "Property": "A124", "Purpose": "A49", "status": "1", "Number of people being liable to provide maintenance for": "1", "Number of existing credits at this bank": "1", "Present employment since": "A75", "Other installment plans": "A143", "Installment rate in percentage of disposable income": "4", "Present residence since": "3", "Status of existing checking account": "A12", "Savings account/bonds": "A61", "Telephone": "A191", "Other debtors / guarantors": "A101", "Age in years": "59", "Duration in month": "48", "Credit history": "A31", "foreign worker": "A201"\}.\textbackslash n Example three: \{"Housing": "A151", "Installment rate in percentage of disposable income": "4", "Age in years": "31", "Duration in month": "24", "foreign worker": "A201", "Number of people being liable to provide maintenance for": "1", "Other installment plans": "A143", "Savings account/bonds": "A61", "Present employment since": "A73", "Credit history": "A31", "Status of existing checking account": "A11", "Job": "A173", "Telephone": "A192", "Number of existing credits at this bank": "1", "status": "1", "Personal status and sex": "A93", "Credit amount": "3161", "Other debtors / guarantors": "A101", "Purpose": "A49", "Property": "A122", "Present residence since": "2"\}.\textbackslash n Example four: \{"Purpose": "A49", "Number of people being liable to provide maintenance for": "1", "Housing": "A151", "Age in years": "26", "Savings account/bonds": "A62", "Other installment plans": "A143", "Present employment since": "A73", "Telephone": "A191", "Installment rate in percentage of disposable income": "4", "Duration in month": "30", "Number of existing credits at this bank": "2", "Personal status and sex": "A92", "Present residence since": "4", "Status of existing checking account": "A12", "Job": "A172", "Credit history": "A30", "Property": "A123", "Other debtors / guarantors": "A101", "status": "1", "Credit amount": "4280", "foreign worker": "A201"\}.\textbackslash n Example five: \{"Present employment since": "A74", "Credit amount": "3566", "Duration in month": "48", "foreign worker": "A201", "Other debtors / guarantors": "A101", "Other installment plans": "A143", "Number of existing credits at this bank": "1", "Number of people being liable to provide maintenance for": "1", "Credit history": "A31", "Housing": "A152", "Present residence since": "2", "Installment rate in percentage of disposable income": "4", "Savings account/bonds": "A62", "Telephone": "A191", "status": "0", "Job": "A173", "Purpose": "A49", "Age in years": "30", "Personal status and sex": "A93", "Property": "A123", "Status of existing checking account": "A12"\}.\textbackslash n Generate one sample: \\
      \midrule
      \fontsize{10}{10}\selectfont \textbf{OUTPUT:} \{"Present residence since": "4", "Credit amount": "7685", "Age in years": "37", "Other installment plans": "A143", "Status of existing checking account": "A11", "Housing": "A151", "Credit history": "A31", "Duration in month": "48", "Property": "A123", "Purpose": "A49", "Other debtors / guarantors": "A103", "Present employment since": "A74", "Installment rate in percentage of disposable income": "2", "Job": "A173", "Savings account/bonds": "A61", "Telephone": "A191", "Number of people being liable to provide maintenance for": "1", "Number of existing credits at this bank": "1", "Personal status and sex": "A92", "foreign worker": "A201", "status": "1"\}.\\
      \bottomrule
    \end{tabular}
  }
  \caption{An instance of the instruction data for the generator}
  \label{a instance of generator}%
\end{table*}

\begin{table*}[!htb]
  \centering
  \resizebox{0.999\textwidth}{!}{
    \begin{tabular}{p{0.999\textwidth}}
      \toprule

       \fontsize{10}{10}\selectfont \textbf{INPUT:} Evaluate the creditworthiness of a customer with the following financial profile. Respond with only either 'good' or 'bad'. \textbackslash n Text: 'The state of Status of existing checking account is bigger than 0 DM but smaller than 200 DM, The state of Duration in month is 36, The state of Credit history is delay in paying off in the past, The state of Purpose is car (new), The state of Credit amount is 1873, The state of Savings account or bonds is bigger than 100 smaller than  500 DM, The state of Present employment since is bigger than 1  smaller than 4 years, The state of Installment rate in percentage of disposable income is 2, The state of Personal status and sex is male and single, The state of  Other debtors or guarantors is none, The state of Present residence since is 2, The state of Property is unknown or no property, The state of Age in years is 29, The state of Other installment plans is none, The state of Housing is for free, The state of Number of existing credits at this bank is 1.0, The state of Job is management or self-employed or highly qualified employee or officer, The state of Number of people being liable to provide maintenance for is 1, The state of Telephone is yes, registered under the customers name, The state of foreign worker is yes.'\textbackslash n Answer: \\
      \midrule
      \fontsize{10}{10}\selectfont \textbf{OUTPUT:} "bad"\\
      \bottomrule
    \end{tabular}
  }
  \caption{An instance of the instruction data for downstream tasks}
  \label{a instance of downstream tasks}%
\end{table*}


\section{Experimental Details}
\label{Exp}

\subsection{Parameter Selection}

Considering the time cost of the entire experiment, we did not adjust the best hyperparameters for different dataset. By conducting experiments on the validation set and combining empirical settings, we unified the hyperparameters of the fine-tuning process. In the fine-tuning stage, we choose lora\cite{hu2021lora} efficient fine-tuning instead of full parameter adjustment.


We fine-tune the LLaMA-2-7b-chat model for each dataset for 5 epochs with a batch size of 16. We utilize the AdamW optimizer for the proposed generative models, with the learning rate $3\times 10^{-4}$.

For the sampling step, we use 3 random seeds in the data generation stage for each dataset, specifically 1234, 1235, and 1236. We set the temperature parameter T to 0.7 for all experiments and datasets. We sample new synthetic data using the prompt dataset for generation (Sec \ref{data pre}), starting with task description and five random real samples(see an example in Appendix \ref{data instance}). We generated synthetic datasets for German and Diabetes with the same number of samples as their respective training sets. For the Adult Income and Buddy datasets, where the training sets are larger, exceeding 10,000 samples, we generated 5,000 samples due to the extended time required for sampling with our method.

For the MLE metric, we employ logistic regression, decision tree, mlp and random forest models.

For the LLE metric, the epoch set for fine-tuning the downstream LLaMA-2-7b-chat model is 5, the learning rate is $1\times 10^{-4}$, and the batch size is 32. The random seed is fixed when fine-tuning the downstream model. See an example of instruction data for downstream tasks in Appendix \ref{data instance}).

\subsection{Experimental Environment}

Our hardware setup includes 4 NVIDIA A100-40GB GPUs. The system has 1 TB system RAM, and runs on an AMD EPYC 7742 processor with 64 cores, using the Ubuntu 22.04 operating system.

\section{Additional results}
\label{additional results}
The following presents the results of the ablation study. We conducted comparative experiments using the German and Diabetes datasets.

\subsection{Filter operation}
Experimental results demonstrate that the filtering step can enhance the quality of synthetic data. As shown in Table \ref{tab:filter}, the LLE values decrease without filtering, particularly for the German dataset. This is likely due to incorrect labels in the generated synthetic data. Additionally, privacy slightly diminishes without the filtering step, though the difference is minimal. These findings indicate that the filtering step is effective.
\renewcommand{\arraystretch}{1.2}
\begin{table*}[!htb]
\centering
\caption{The results of whether to filter data after kNN, where "w/o fil" means not to filter data, and "with fil" means to filter data, which is our original method. Each dataset has five metrics, and in all cases, higher values are better.}
\vspace{1mm}
\label{tab:filter}
\resizebox{0.999\textwidth}{!}{
    \begin{tabular}{
    >{\centering\arraybackslash}m{1.3cm}
    >{\centering\arraybackslash}m{1.5cm}
    >{\centering\arraybackslash}m{1.3cm}
    >{\centering\arraybackslash}m{1.3cm}
    >{\centering\arraybackslash}m{1.3cm}
    >{\centering\arraybackslash}m{1.3cm}
    >{\centering\arraybackslash}m{1.3cm}}
    \toprule
    Dataset & Filter & MLE & LLE & NRS & DCR & DLT \\
    \midrule
    \multirow{2}{*}{GM} & w/o fil & $0.56_{\pm 0.06}$ & $0.59_{\pm 0.03}$ & 1.00 & 7.97 & -0.17 \\
     & with fil & $0.55_{\pm 0.03}$ & $0.64_{\pm 0.03}$ & 1.00 & 8.08 & -0.16 \\

    \midrule
    \multirow{2}{*}{DI} & w/o fil & $0.56_{\pm 0.06}$ & $0.74_{\pm 0.01}$ & 1.00 & 0.44 & -0.38 \\
     & with fil & $0.46_{\pm 0.02}$ & $0.75_{\pm 0.00}$ & 1.00 & 0.44 & -0.37 \\
    
    \bottomrule
    \end{tabular}
}
\vspace{-3mm}
\end{table*}

\subsection{Random feature order permutation}
Experiments indicate that permuting features can enhance the privacy of synthetic data. As shown in the last two columns of Table \ref{tab:permutation}, there is a significant reduction in both the DCR and DLT values when features are not permuted. Concurrently, the generated numerical columns tend to produce repeated values, which may also contribute to the decrease in the LLE metric. Overall, these results underscore the necessity of shuffling features.
\renewcommand{\arraystretch}{1.2}
\begin{table*}[!htb]
\centering
\caption{The results of whether to shuffle features, where "w/o pm" means not to shuffle the features, and "with pm" means to shuffle the features, which is our original method. Each dataset has five metrics, and in all cases, higher values are better.}
\vspace{1mm}
\label{tab:permutation}
\resizebox{0.999\textwidth}{!}{
    \begin{tabular}{
    >{\centering\arraybackslash}m{1.3cm}
    >{\centering\arraybackslash}m{1.5cm}
    >{\centering\arraybackslash}m{1.3cm}
    >{\centering\arraybackslash}m{1.3cm}
    >{\centering\arraybackslash}m{1.3cm}
    >{\centering\arraybackslash}m{1.3cm}
    >{\centering\arraybackslash}m{1.3cm}}
    \toprule
    Dataset & Permutation & MLE & LLE & NRS & DCR & DLT \\
    \midrule
    \multirow{2}{*}{GM} & w/o pm & $0.56_{\pm 0.04}$ & $0.63_{\pm 0.05}$ & 1.00 & 7.20 & -0.58\\
     & with pm & $0.55_{\pm 0.03}$ & $0.64_{\pm 0.03}$ & 1.00 & 8.08 & -0.16 \\

    \midrule
    \multirow{2}{*}{DI} & w/o pm & $0.50_{\pm 0.06}$ & $0.70_{\pm 0.03}$ & 1.00 & 0.42 & -0.67 \\
     & with pm & $0.46_{\pm 0.02}$ & $0.75_{\pm 0.00}$ & 1.00 & 0.44 & -0.37 \\
    
    \bottomrule
    \end{tabular}
}
\vspace{-3mm}
\end{table*}

\section{Ethics Statement}
\label{eth}

The dataset used in this study is based on open-source data and can be further modified. We thoroughly reviewed and verified the data to ensure it does not contain any personally identifiable information or offensive content. Additionally, we conducted manual audits to ensure there are no sensitive details. Therefore, we believe the dataset is secure and its use in the research is ethically sound and appropriate for the purposes of this study.


\end{document}